\documentclass[letterpaper, 10 pt, conference]{ieeeconf}  
\usepackage[left=0.75in,right=0.75in,top=0.75in,bottom=0.75in]{geometry}

\IEEEoverridecommandlockouts                              

\usepackage[dvipdfmx]{graphicx}
\usepackage{amsmath}
\usepackage{comment}
\usepackage{url}
\usepackage{color}
\usepackage{subfigure}
\usepackage{mathtools}
\usepackage{cite}
\usepackage{algorithm}
\usepackage{algorithmic}
\usepackage{multirow}
\usepackage{amsfonts}
\usepackage{amssymb}
\usepackage{balance}

\title{\LARGE \bf
LiDAR-Inertial Odometry Based on Extended Kalman Filter
}

\author{Naoki Akai$^{1,2}$ and Takumi Nakao$^{3}$
\thanks{$^{1}$Naoki Akai is with the Institute of Innovation for Future Society, Nagoya University, Nagoya 464-8603, Japan {\tt\small akai@nagoya-u.jp}}%
\thanks{$^{2}$Naoki Akai is with the LOCT Co., Ltd., Nagoya 466-0805, Japan}%
\thanks{$^{3}$Takumi Nakao is with the Graduate School of Science and Technology, University of Tsukuba, Japan}%
}

\newcommand{\argmin}{\mathop{\rm argmin}\limits}

\begin{document}

\newcommand{\1}{\mbox{1}\hspace{-0.25em}\mbox{l}}
\renewcommand{\baselinestretch}{1.0}

\maketitle
\thispagestyle{empty}
\pagestyle{empty}

\begin{abstract}

LiDAR-Inertial Odometry (LIO) is typically implemented using an optimization-based approach, with the factor graph often being employed due to its capability to seamlessly integrate residuals from both LiDAR and IMU measurements. Conversely, a recent study has demonstrated that accurate LIO can also be achieved using a loosely-coupled method. Inspired by this advancements, we present a LIO method that leverages the recursive Bayes filter, solved via the Extended Kalman Filter (EKF) - herein referred to as KLIO. Within KLIO, prior and likelihood distributions are computed using IMU preintegration and scan matching between LiDAR and local map point clouds, and the pose, velocity, and IMU biases are updated through the EKF process. Through experiments with the Newer College dataset, we demonstrate that KLIO achieves precise trajectory tracking and mapping. Its accuracy is comparable to that of the state-of-the-art methods in both tightly- and loosely-coupled methods.

\end{abstract}


\section{Introduction}

Estimating continuous ego-motion is a crucial functionality for mobile robots, serving as the foundation for their autonomous navigation capabilities.
Traditionally, odometry systems have been implemented using wheel encoders and/or gyroscopic sensors, known as wheel and/or gyro odometry~\cite{BorensteinJRS1997}.
While these systems are effective, their applicability is limited to wheeled mobile robots due to the reliance on wheel encoders.
The recent advancements in 3D LiDAR technology have paved the way for estimating ego-motion by matching continuous measurements, a method known as LiDAR Odometry (LO) and/or LiDAR Odometry and Mapping (LOAM)~\cite{loam2014zhang, legoloam2018shan, chen2022direct}.
LO is capable of estimating accurate ego-motion but struggles to track rapid movements, especially rotations, due to the low measurement frequency of LiDAR.
To address this limitation, the integration of LO with IMU, referred to as LiDAR-Inertial Odometry (LIO), has been proposed by numerous researchers recently.
This approach leverages the strengths of both LiDAR and IMU measurements to achieve more robust and accurate ego-motion estimation, especially in scenarios involving quick movements or rotations.

LIO is typically implemented through a tightly-coupled method, optimizing a cost function that incorporates residuals from both LiDAR and IMU measurements~\cite{LIO-mapping, liosam2020shan, DingICRA2020, DongRA-L2023}.
This optimization process not only estimates sensor pose but also corrects for velocity and IMU biases, leading to precise pose estimation and rapid tracking of ego-motion.
Xu {\it et al.}~\cite{FAST-LIO, FAST-LIO2} introduced FAST-LIO and FAST-LIO2, which utilize an iterated Extended Kalman Filter (EKF) for their implementation.
These methods focus on optimizing the residuals from LiDAR measurements while also estimating IMU biases through iterative updates.
In contrast, Chen {\it et al.}~\cite{Direct-LIO} proposed a loosely-coupled LIO approach referred to as Direct LIO (DLIO), which has demonstrated superior performance over other LIO methods.
DLIO leverages a hierarchical geometric observer~\cite{Lopez2023arXiv} to update the estimated state, including sensor pose, velocity, and IMU biases.
These updates are derived from the pose differences between prediction and estimate, with the latter obtained through the scan matching process between LiDAR and local map point clouds.
Consequently, DLIO is classified as a loosely-coupled method.

\begin{figure*}[!t]
  \begin{center}
    \includegraphics[width = 175 mm]{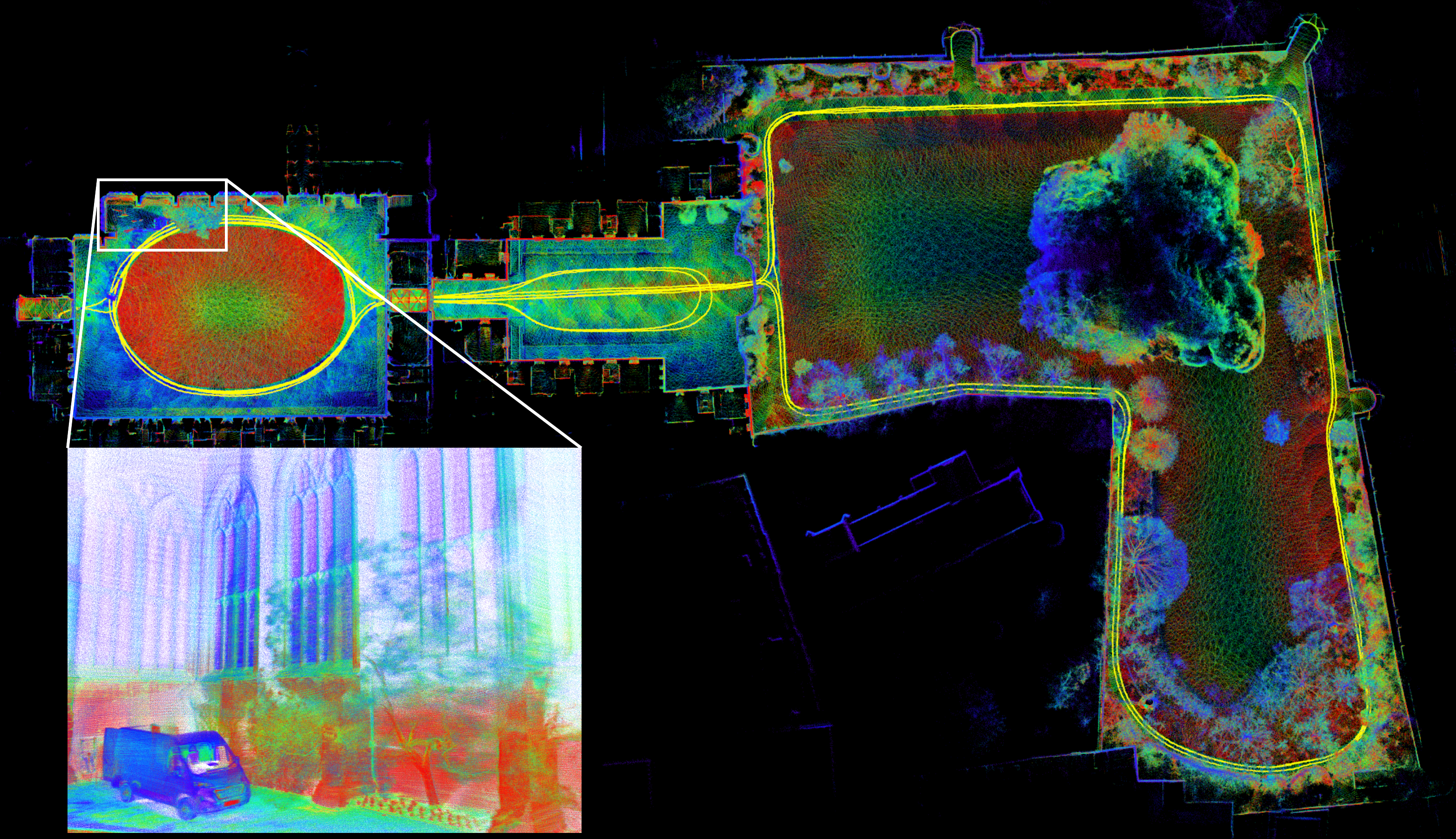}
    \caption{A mapping result using KLIO on the {\rm 02\_long\_experiment} from the Stereo-Cam dataset, part of the Newer College dataset~\cite{ramezani2020newer}, demonstrates KLIO's effectiveness. Despite the absence of a loop-closure process, it successfully closes large loops, with the total walking distance exceeding 3~km. The estimated trajectory is illustrated by the yellow line, and the colors of the points reflect the LiDAR intensity. The bottom left detailed view reveals the precise reconstruction of cars, trees, and windows.}
    \label{fig:result_ncd_long}
  \end{center}
\end{figure*}

Inspired by this work, we present a novel LIO method that utilizes the recursive Bayes filter, resolved through the Extended Kalman Filter (EKF), herein referred to as KLIO.
While KLIO shares similarities with DLIO, it distinguishes itself through a state update process grounded in EKF principles.
In KLIO, prediction is performed using IMU preintegration~\cite{ForsterTRO2017}, and measurements for the update phase are derived from scan matching results between LiDAR and local map point clouds.
Despite its straightforward system architecture, KLIO achieves high performance.
An example mapping result using KLIO on the 02\_long\_experiment from the Stereo-Cam dataset within the Newer College dataset~\cite{ramezani2020newer} is shown in Fig.~\ref{fig:result_ncd_long}.
This scenario, spanning a considerable area with a total walking distance of over 3~km, demonstrates KLIO's ability to accurately close loops, even in the absence of a loop-closure function.
In addition, we reveal robustness of KLIO trough an experiment involving rapid rotation exceeding $4~{\rm rad/sec}$.

The rest of this paper is organized follow.
Section~\ref{sec:related_work} summarizes related works.
Section~\ref{sec:proposed_method} details the proposed method.
Section~\ref{sec:experiments} describes experimental results.
Section~\ref{sec:conclusion} concludes this work.

\section{Related work}
\label{sec:related_work}

The principal functionality of LO and LIO is point cloud registration.
Predominant methods in point cloud registration include the Iterative Closest Point (ICP)~\cite{BeslTPAMI1992} and the Normal Distributions Transform (NDT)~\cite{biber_iros2003:_ndt}.
These methods are classified as point-to-point and point-to-distribution matching techniques.
Although these techniques are capable of utilizing the entirety of point sets, this may lead to an escalation in computational cost.
Fature-based registration approaches such as edges, lines, and planes have been introduced to increase computational efficiency~\cite{loam2014zhang, legoloam2018shan, FAST-LIO, FAST-LIO2, Li2020TowardsHS, Huang2023LOGLIOAL}.

Recent advancements in 3D LiDAR technology have made it possible to capture a vast number of points in a single measurement.
The Generalized ICP (GICP)~\cite{SegalRSS2009GICP}, which accounts for the shapes of point clouds from both the source and target by considering their point distributions, has been shown to work efficiently and accurately.
Consequently, GICP-based LIO and/or SLAM methods have been introduced in recent~\cite{chen2022direct, Direct-LIO, FastGICP, KoideICRA2022}.
Extensions of GICP algorithms can be seen in~\cite{multi-channel-GICP, Surface-GICP}.
Yokozuka {\it et al}.~\cite{LiTAMIN} introduced a cost function that normalizes the covariance matrix used in the GICP cost function.
Furthermore, Yokozuka {\it et al}.~\cite{LiTAMIN2} proposed a cost function inspired by the Kullback-Leibler divergence.
These methods including GICP are classified as distribution-to-distribution matching techniques and have exhibited rapid, robust, and precise registration capabilities.
In our research, we adopt GICP for scan matching purposes.

LO is adversely affected by rapid movements, particularly rotations.
To mitigate this issue, fusion with an IMU is employed.
Ye {\it et al}.~\cite{LIO-mapping} and Shan {\it et al}.~\cite{liosam2020shan} have introduced a tightly-coupled LIO approach that optimizes residuals using data from both LiDAR and IMU.
These approaches leverage a factor graph framework~\cite{FactorGraph}.
Xu {\it et al}.~\cite{FAST-LIO, FAST-LIO2} developed FAST-LIO and FAST-LIO2, which, unlike previous methods, do not optimize residuals based on IMU data directly.
Instead, they allow for the simultaneous optimization of velocity and IMU biases through an iterated extended Kalman filter~\cite{huai2023quick}.
As such, these methods are classified as tightly-coupled approaches.
The findings from these studies underscore the necessity of tight coupling for achieving accurate LIO outcomes.

However, Chen {\it et al}.~\cite{Direct-LIO} introduced DLIO, demonstrating that accurate LIO can be achieved through a loosely-coupled approach.
In DLIO, the initial step involves scan matching between the LiDAR and local map point clouds. Subsequently, the pose, velocity, and IMU biases are updated based on the results of the scan matching.
This update mechanism is referred to as a hierarchical geometric observer~\cite{Lopez2023arXiv}, which adjusts the state by accounting for the discrepancies between the predicted and estimated poses, as determined through IMU preintegration and scan matching.

Our work is inspired by DLIO and introduces a novel probabilistic approach to LIO.
Our approach formulates the LIO problem within the framework of the recursive Bayes filter and solves it using EKF.
We posit that the outcomes of scan matching serve as observations for the purpose of updating a predicted state, a process akin to that utilized by the geometric observer.
Similar approach is recently presented in~\cite{wu2024icra}.
However, our method diverges by assuming that observations pertinent to velocity and IMU biases can also be derived from the results of two consecutive scan matching results, whilst incorporating IMU preintegration.
Consequently, we believe our approach offers enhanced robustness compared to DLIO.

\section{Proposed method}
\label{sec:proposed_method}

\subsection{Problem setting and proposed model}

In this work, we address the LIO problem, which involves estimating the current state using LiDAR and IMU measurements, denoted as $\mathcal{P}$ and ${\bf u} \in \mathbb{R}^{6}$, respectively.
We introduce a method to solve this problem using EKF and refer to our proposal as KLIO (LiDAR-Inertial Odometry based on Extended Kalman Filter).

The target state for the estimate is denoted as
%
\begin{align}
  \begin{gathered}
    {\bf x}_{t} = 
    [ {}^{\rm w}{\bf R}_{t} ~~ {}^{\rm w}{\bf p}_{t} ~~ {}^{\rm w}{\bf v}_{t}
    ~~ {}^{\rm \omega}{\bf b}_{t} ~~ {}^{\rm a}{\bf b}_{t} ],
  \end{gathered}
  \label{eq:target_state}
\end{align}
where ${}^{\rm w}{\bf R}_{t} \in SO(3)$ represents the rotation matrix, ${}^{\rm w}{\bf p}_{t} \in \mathbb{R}^{3}$ the translation vector, and ${}^{\rm w}{\bf v}_{t} \in \mathbb{R}^{3}$ the velocity vector of the IMU in world coordinates.
Additionally, ${}^{\rm \omega}{\bf b}_{t} \in \mathbb{R}^{3}$ and ${}^{\rm a}{\bf b}_{t} \in \mathbb{R}^{3}$ represent the IMU biases for gyroscope and acceleration measurements, respectively.
For ease of notation, we express the pose using the homogeneous matrix ${}^{\rm w}{\bf T}_{t} = ({}^{\rm w}{\bf R}_{t} | {}^{\rm w}{\bf p}_{t}) \in SE(3)$.
LIO constructs a local point cloud map $\mathcal{M}$ by aggregating LiDAR measurements.
In our approach, scan matching is executed against this local map, and its outcome is utilized as an observation in the EKF update phase.
This observation is denoted as
\begin{align}
  {\bf z}_{t} = 
  [ {}^{\rm w}\tilde{{\bf R}}_{t} ~~ {}^{\rm w}\tilde{{\bf p}}_{t}
  ~~ {}^{\rm w}\tilde{{\bf v}}_{t}
  ~~ {}^{\rm \omega}\tilde{{\bf b}}_{t} ~~ {}^{\rm a}\tilde{{\bf b}}_{t} ],
  \label{eq:measurement}
\end{align}
where ${}^{\rm w}\tilde{{\bf T}}_{t} = ( {}^{\rm w}\tilde{{\bf R}}_{t} | {}^{\rm w}\tilde{{\bf p}}_{t} )$ represents the scan matching result, i.e., the optimal estimate at time $t$, and ${}^{\rm w}\tilde{{\bf v}}_{t}$, ${}^{\rm \omega}\tilde{{\bf b}}_{t}$, and ${}^{\rm a}\tilde{{\bf b}}_{t}$ are derived from two consecutive scan matching results.
These measurements are elaborated upon in Section~\ref{subsec:update}.
The IMU measurements gathered between the timestamps of consecutive LiDAR measurements, $\rho_{t-1}$ and $\rho_{t}$, are denoted as
\begin{align}
  {\bf u}_{t} = [ \boldsymbol \omega_{t}^{1} ~~ {\bf a}_{t}^{1} ~~ \cdots ~~
  \boldsymbol \omega_{t}^{K} ~~ {\bf a}_{t}^{K} ] \in \mathbb{R}^{6K},
  \label{eq:imu_measurement}
\end{align}
where $\boldsymbol \omega_{t}^{k} \in \mathbb{R}^{3}$ and ${\bf a}_{t}^{k} \in \mathbb{R}^{3}$ are gyro scope and acceleration measurements at time $\rho_{t-1}^{k}$, where $\rho_{t-1} \leq \rho_{t-1}^{k} \leq \rho_{t}$.

The proposed method is formulated as an input-output hidden Markov model~\cite{AkaiIV2019}.
This model estimates the posterior distribution
\begin{align}
  \begin{split}
    & p({\bf x}_{t} | {\bf u}_{1:t}, {\bf z}_{1:t}) = \eta p({\bf z}_{t} | {\bf x}_{t}) \\
    & \int p({\bf x}_{t} | {\bf x}_{t-1}, {\bf u}_{t} )
      p({\bf x}_{t-1} | {\bf u}_{1:t-1}, {\bf z}_{1:t-1}) {\rm d} {\bf x}_{t-1},
  \end{split}
  \label{eq:target_distribution}
\end{align}
where $\eta$ is a normalization constant.
The likelihood distribution $p({\bf z}_{t} | {\bf x}_{t})$ and the prior distribution $\int p({\bf x}_{t} | {\bf x}_{t-1}, {\bf u}_{t} ) p({\bf x}_{t-1} | {\bf u}_{1:t-1}, {\bf z}_{1:t-1}) {\rm d} {\bf x}_{t-1}$ are estimated through scan matching and IMU preintegration, respectively.

\begin{figure*}[!t]
  \begin{center}
    \includegraphics[width = 140 mm]{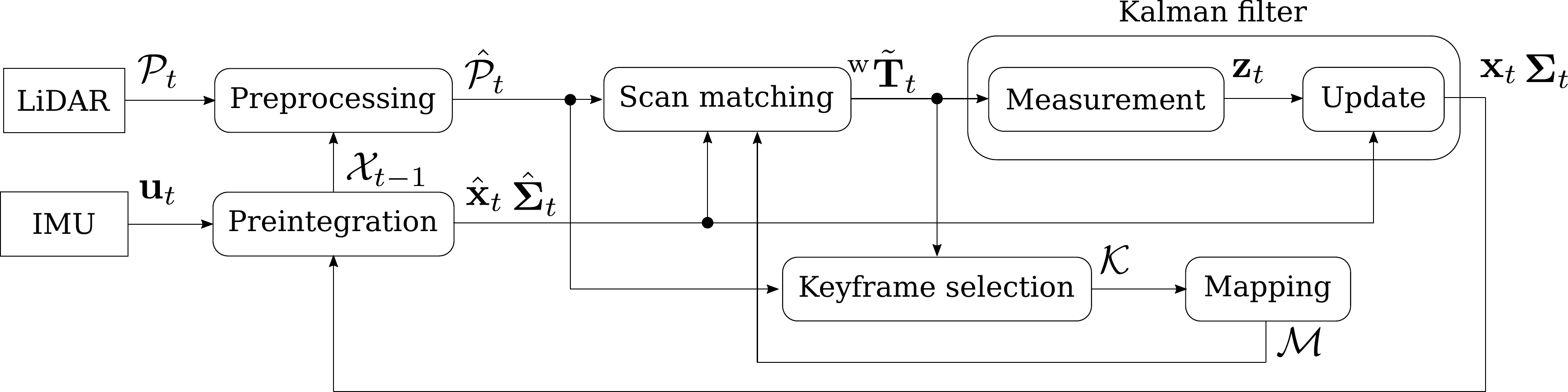}
    \caption{System overview of LiDAR-Inertial Odometry based on Extended Kalman Filter (KLIO).}
    \label{fig:system_overview}
  \end{center}
\end{figure*}

The system overview and pseudocode are depicted in Fig.~\ref{fig:system_overview} and Algorithm~\ref{alg:overview}, respectively.
The subsequent subsections provide details of the processes illustrated in Fig.~\ref{fig:system_overview} and Algorithm~\ref{alg:overview}.
The pseudocode is described using the callbacks because we used ROS to implement source code used in this work and the IMU measurement buffer $\mathcal{U}_{t}$ is shared within the callbacks.

\begin{algorithm}[!t]
  \caption{KLIO}
  \begin{algorithmic}[1]
    \REQUIRE ${\bf x}_{t-1}, \boldsymbol \Sigma_{t-1}, \boldsymbol \omega_{t}^{k}, {\bf a}_{t}^{k}, \mathcal{P}_{t}, \mathcal{M}, \mathcal{K}$
    \ENSURE ${\bf x}_{t}, \boldsymbol \Sigma_{t}, \mathcal{M}, \mathcal{K}$
    \newline \newline
    \textcolor{black}{{\rm // LiDAR callback}}
    \WHILE{$\mathcal{P}_{t} \neq \emptyset$}
      \STATE $\mathcal{T}_{t} \leftarrow$ get\_scan\_points\_timestamps($\mathcal{P}_{t}$)
      \STATE ${\bf u}_{t} \leftarrow$ get\_corresponding\_imu\_measurements($\mathcal{U}_{t}, \mathcal{T}_{t}$)
      \STATE $\mathcal{X}_{t-1}, \hat{ \boldsymbol \Sigma }_{t} \leftarrow$ imu\_preintegration(${}^{\rm w}{\bf T}_{t-1}, {\bf u}_{t}$)
%
%
      \STATE $\hat{ \mathcal{P} }_{t} \leftarrow$ preprocess($\mathcal{P}_{t}, \mathcal{X}_{t-1}$)
      \STATE ${}^{\rm w}\tilde{{\bf T}}_{t}, \tilde{{\bf H}}_{t}, \gamma_{t} =$ scan\_matching(${}^{\rm w}\hat{{\bf T}}_{t}, \hat{ \mathcal{P} }_{t}, \mathcal{M}$)
      \STATE ${\bf z}_{t} \leftarrow$ calculate\_measurement(${}^{\rm w}\tilde{{\bf T}}_{t}, {}^{\rm w}\tilde{{\bf T}}_{t-1}$)
      \STATE ${\bf x}_{t}, \boldsymbol \Sigma_{t} \leftarrow$ kalman\_filter\_update($\hat{{\bf x}}_{t}, \hat{ \boldsymbol \Sigma }_{t}, {\bf z}_{t}, \tilde{{\bf H}}_{t}$)
      \IF{$\gamma_{t} < \gamma_{\rm th}$}
        \STATE $\mathcal{K} \leftarrow \mathcal{K} + ({}^{\rm w}{\bf T}_{t}, \hat{ \mathcal{P} }_{t}$) \textcolor{black}{{\rm // Update keyframe set}}
        \STATE $\mathcal{M} \leftarrow$ build\_new\_local\_map($\mathcal{K}$)
      \ENDIF
    \ENDWHILE
    \newline \newline
    \textcolor{black}{{\rm // IMU callback}}
    \WHILE{$\boldsymbol \omega_{t}^{k} \neq \emptyset$ and ${\bf a}_{t}^{k} \neq \emptyset$}
      \STATE ${\bf u}_{t}^{k} = ( \boldsymbol \omega_{t}^{k}, {\bf a}_{t}^{k} )$
      \STATE $\mathcal{U}_{t} \leftarrow \mathcal{U}_{t} \oplus {\bf u}_{t}^{k}$ \textcolor{black}{{\rm // Add to circular buffer}}
    \ENDWHILE
  \end{algorithmic}
  \label{alg:overview}
\end{algorithm}

It should be noted that the LIO problem can be formulated as a SLAM problem, which entails the joint posterior estimation of the current state and the local map, as the local map is concurrently constructed during the state estimation process.
However, we simplify the problem by positing that LIO is capable of generating an accurate local map by plotting LiDAR measurements according to the optimal estimates.
This assumption allows us to disregard mapping uncertainty.
Through experiments, we demonstrate that precise trajectory estimation and map construction can be achieved by estimating Eq.~(\ref{eq:target_distribution}).

\subsection{Prediction}
\label{subsec:prediction}

To estimate the prior as shown in Eq.~(\ref{eq:target_distribution}), we utilize IMU preintegration~\cite{ForsterTRO2017}.
The propagation model ${\bf f}$, which transitions the state from ${\bf x}_{t-1}^{k-1}$ to ${\bf x}_{t-1}^{k}$ using the IMU measurement ${\bf u}_{t}^{k}$, is expressed as
%
\begin{align}
  \begin{split}
    & {\bf f}({\bf x}_{t-1}^{k-1}, {\bf u}_{t}^{k}, {\bf w}_{t}^{k}) \\
    & = \left[ \begin{matrix}
      {}^{\rm w}{\bf R}_{t-1}^{k-1} \exp( \hat{\boldsymbol \omega}_{t} \Delta t_{k} ) \exp \left( \frac{1}{2} \boldsymbol \alpha_{t}^{k} \Delta t_{k}^{2} \right) \\
      {}^{\rm w}{\bf p}_{t-1}^{k-1} + \Delta {}^{\rm w}{\bf p}_{t-1}^{k-1} \\
      {}^{\rm w}{\bf v}_{t-1}^{k-1} + {\bf g} \Delta t_{k} + {}^{\rm w}{\bf R}_{t-1}^{k-1} ( \hat{{\bf a}}_{t}^{k} ) \Delta t_{k} + \frac{1}{2} {\bf j}_{t}^{k} \Delta t_{k}^{2} \\
      {}^{\rm \omega}{\bf b}_{t-1}^{k-1} + {}^{\rm \omega}{\bf m}_{t}^{k} \\
      {}^{\rm a}{\bf b}_{t-1}^{k-1} + {}^{\rm a}{\bf m}_{t}^{k} \\
    \end{matrix} \right]
  \end{split}
\end{align}
\begin{align}
  \begin{split}
    \Delta {}^{\rm w}{\bf p}_{t-1}^{k-1} = & {}^{\rm w}{\bf v}_{t-1}^{k-1} \Delta t_{k} + \\
    & \frac{1}{2} {\bf g} \Delta t_{k}^{2} + \frac{1}{2} {}^{\rm w}{\bf R}_{t-1}^{k-1} ( \hat{{\bf a}}_{t}^{k} ) \Delta t_{k}^{2} + \frac{1}{6} {\bf j}_{t}^{k} \Delta t_{k}^{3}
  \end{split}
\end{align}
where $\Delta t_{k}$ denotes the time difference between $\rho_{t-1}^{k-1}$ and $\rho_{t-1}^{k}$, ${\bf g}$ represents the gravity vector, and ${\bf w}_{t}^{k} = [ {}^{\rm \omega}{\bf n}_{t}^{k} \quad {}^{\rm a}{\bf n}_{t}^{k} \quad {}^{\rm \omega}{\bf m}_{t}^{k} \quad {}^{\rm a}{\bf m}_{t}^{k} ]$ is the noise vector for IMU measurements and biases.
The corrected angular velocity and acceleration are given by $\hat{\boldsymbol \omega}_{t} = \boldsymbol \omega_{t}^{k} - {}^{\rm \omega}{\bf b}_{t}^{k-1} - {}^{\rm \omega}{\bf n}_{t}^{k}$ and $\hat{\bf a}_{t} = {\bf a}_{t}^{k} - {}^{\rm a}{\bf b}_{t}^{k-1} - {}^{\rm a}{\bf n}_{t}^{k}$, respectively.
Additionally, $\boldsymbol \alpha_{t}^{k}$ and ${\bf j}_{t}^{k}$ are angular acceleration and linear jerk denoted as
\begin{align}
  \begin{gathered}
    \boldsymbol \alpha_{t}^{k} = \frac{1}{\Delta t_{k}} \left( \hat{ \boldsymbol \omega }_{t}^{k} - \hat{ \boldsymbol \omega }_{t}^{k-1} \right), \\
    {\bf j}_{t}^{k} = \frac{1}{\Delta t_{k}} \left( {}^{\rm w}{\bf R}_{t-1}^{k-1} ( \hat{{\bf a}}_{t}^{k} ) - {}^{\rm w}{\bf R}_{t-1}^{k-2} ( \hat{{\bf a}}_{t}^{k-1} )  \right).
  \end{gathered}
\end{align}
IMU preintegration yields a set of $K+1$ states, denoted as $\mathcal{X}_{t-1} = [{\bf x}_{t-1}^{0} \quad {\bf x}_{t-1}^{1} \quad \cdots \quad {\bf x}_{t-1}^{K}]$, where ${\bf x}_{t-1}^{0}$ is the previously estimated state and ${\bf x}_{t-1}^{K}$ serves as the predicted state at time $t$, denoted as $\hat{{\bf x}}_{t}$.
During the propagation process, it is assumed that the noise vector equals zero and the state is propagated via ${\bf f}({\bf x}_{t-1}^{k-1}, {\bf u}_{t}^{k}, {\bf 0})$.

The covariance of the state, denoted by $\boldsymbol \Sigma$, is also propagated using IMU preintegration.
The partial derivatives of ${\bf f}$ with respect to ${\bf x}$ and ${\bf w}$ are denoted as
\begin{align}
  \begin{gathered}
    \left. \frac{ \partial {\bf f}( {\bf x}_{t-1}^{k-1}, {\bf u}_{t}^{k}, {\bf w}_{t}^{k} ) }{ \partial {\bf x} } \right|_{ {\bf x} = {\bf x}_{t-1}^{k-1} } = {\bf F}_{\bf x}, \\
    \left. \frac{ \partial {\bf f}( {\bf x}_{t-1}^{k-1}, {\bf u}_{t}^{k}, {\bf w}_{t}^{k} ) }{ \partial {\bf w} } \right|_{ {\bf w} = {\bf 0} } = {\bf F}_{\bf w},
  \end{gathered}
  \label{eq:preintegration_derivatives}
\end{align}
the covariance is then propagated as
\begin{align}
  \boldsymbol \Sigma_{t-1}^{k} = {\bf F}_{\bf x} \boldsymbol \Sigma_{t-1}^{k-1} {\bf F}_{\bf x}^{\top} + {\bf F}_{\bf w} {\bf Q} {\bf F}_{\bf w}^{\top},
\end{align}
where ${\bf Q}$ represents the covariance matrix associated with ${\bf w}$, defined as ${\bf Q} = q {\bf I}$ in implementation, with $q$ being a positive arbitrary constant and ${\bf I}$ is the identity matrix.
The covariance $\boldsymbol \Sigma_{t-1}^{K}$ is utilized as the predicted covariance, denoted as $\hat{\boldsymbol \Sigma}_{t}$.
Note that we referred \cite{FAST-LIO} to derive the derivatives shown in Eq.~(\ref{eq:preintegration_derivatives}).

\subsection{Scan matching}
\label{subsec:scan_matching}

After the prediction phase, scan matching is executed against the local map $\mathcal{M}$.
The LiDAR scan points $\mathcal{P}_{t}$ are first deskewed based on the result of IMU preintegration, and a voxel grid filter with a resolution of $r$ is applied to the deskewed points to reduce the size of the point cloud.
It should be noted that the LiDAR point cloud is transformed into IMU coordinates to simplify the fusion of the preintegration results with scan matching estimates, which specifically aim to determine the pose of the IMU.

To deskew the scan points, the preintegration result closest in timestamp to each LiDAR scan point is identified.
Each scan point is then transformed according to the corresponding preintegration result, taking into account the time difference between the scan points and preintegration.
Subsequently, the transformed points are re-transformed to the coordinates of the predicted IMU pose.
The deskewing process modifies the $i$th LiDAR scan point ${\bf p}_{i} \in \mathcal{P}$ as
\begin{align}
  \hat{{\bf p}}_{i} = {}^{\rm w}\hat{{\bf T}}_{t}^{-1} {}^{\rm w}\breve{\bf T}_{t-1}^{k} {\bf p}_{i} \in \mathcal{ \hat{P} }
\end{align}
where ${}^{\rm w}\hat{{\bf T}}_{t}$ represents the predicted IMU pose at time $t$, and ${}^{\rm w}\breve{\bf T}_{t-1}^{k}$ denotes the IMU pose at a specific preintegration step $k$, adjusted to account for the time difference between the IMU pose ${}^{\rm w}{\bf T}_{t-1}^{k}$ and the timestamp of the LiDAR scan point ${\bf p}_{i}$.

For scan matching, we utilize GICP~\cite{SegalRSS2009GICP}.
GICP aims to minimize the following cost function:
\begin{align}
  \begin{gathered}
    {}^{\rm w}\tilde{{\bf T}}_{t} = \argmin_{ {}^{\rm w}{\bf T}_{t} } \sum_{i} {\bf d}_{i}^{\top} \left( {\bf C}_{q} + {}^{\rm w}{\bf T}_{t} {\bf C}_{p} {}^{\rm w}{\bf T}_{t}^{\top} \right)^{-1} {\bf d}_{i}, \\
    {\bf d}_{i} = {\bf q}_{i} - {}^{\bf w}{\bf T}_{t} {\bf p}_{i}, \\
  \end{gathered}
\end{align}
where ${\bf q} \in \mathcal{M}$ represents the point corresponding to ${\bf p} \in \mathcal{P}_{t}$, and ${\bf C}_{p}$ and ${\bf C}_{q}$ are the covariances calculated using the points surrounding ${\bf p}$ and ${\bf q}$, respectively.
The optimization is carried out using the Gauss-Newton method, and the Hessian matrix derived in the final optimization step is denoted as $\hat{{\bf H}}_{t}$.
This Hessian matrix is utilized in the subsequent update phase to ascertain the measurement uncertainty.
Additionally, the correspondence rate $\gamma_{t}$ between the scan points and the local map is computed to determine the appropriate timing for updating the local map.
Only correspondences that satisfy $\| {\bf q} - {}^{\bf w}{\bf T}_{t} {\bf p} \|_{2} \leq \epsilon$ are considered in the optimization process, where $\epsilon$ is a predefined positive constant.

\subsection{Update}
\label{subsec:update}

State update is executed using EKF subsequent to scan matching.
Observations related to the rotation and translation components of the IMU pose are derived from the scan matching result, denoted as ${}^{\rm w}\tilde{{\bf T}}_{t}$.
Additional observations, specifically those concerning velocity and IMU measurement biases, are determined based on two consecutive scan matching results, taking into account the IMU preintegration propagation model:
\begin{align}
  \begin{gathered}
    {}^{\rm w}\tilde{{\bf v}}_{t} = \frac{ {}^{\rm w}\tilde{{\bf p}}_{t} - {}^{\rm w}\tilde{{\bf p}}_{t-1} }{ \Delta t }, \\
    {}^{\rm \omega}\tilde{{\bf b}}_{t} = \bar{\boldsymbol \omega}_{t} - \frac{1}{\Delta t} \log \left( {}^{\rm w}\tilde{{\bf R}}_{t-1}^{\top} {}^{\rm w}\tilde{{\bf R}}_{t} \right), \\
    {}^{\rm a}\tilde{{\bf b}}_{t} = \bar{\bf a}_{t} - \frac{1}{\Delta t} \tilde{{\bf R}}_{t-1}^{\top} \left( {}^{\rm w}\tilde{{\bf v}}_{t} - {}^{\rm w}\tilde{{\bf v}}_{t-1} - {\bf g}\Delta t \right),
  \end{gathered}
  \label{eq:observation}
\end{align}
where $\Delta t = \rho_{t} - \rho_{t-1}$ represents the time interval between two consecutive LiDAR measurements, and $\bar{\boldsymbol \omega}_{t}$ and $\bar{\bf a}_{t}$ denote the average gyroscope and acceleration measurements within ${\bf u}_{t}$, respectively.
In Eq.~(\ref{eq:observation}), it is assumed that the translation and angular velocities can be inferred from two consecutive scan matching results, and that the IMU biases can be derived from these velocities in conjunction with the average IMU measurements.

Then, state update is performed as
\begin{align}
  \begin{gathered}
    {\bf x}_{t} = \hat{{\bf x}}_{t} \boxplus {\bf K}_{t} ( {\bf z}_{t} \boxminus \hat{{\bf x}}_{t} ) ~~~~~
    \boldsymbol \Sigma_{t} = ( {\bf I} - {\bf K}_{t} ) \hat{ \boldsymbol \Sigma }_{t}, \\
    {\bf K}_{t} = \hat{ \boldsymbol \Sigma }_{t} \left( {\bf R}_{t} + \hat{ \boldsymbol \Sigma }_{t} \right)^{-1}, \\
  \end{gathered}
\end{align}
where ${\bf I}$ is the identity matrix, and ${\bf R}_{t}$ represents the covariance matrix that encapsulates the uncertainty associated with ${\bf z}_{t}$.
Ascertaining precise values for ${\bf R}_{t}$ poses a significant challenge.
To estimate the uncertainty pertaining to the pose, we reference~\cite{BengtssonIROS2001}, which suggests utilizing the Hessian matrix derived during the optimization process for this purpose.
The optimization is conducted via the Gauss-Newton method, enabling the acquisition of the Hessian matrix at the final optimization stage, denoted as $\tilde{{\bf H}}_{t}$.
Nevertheless, additional indicators for determining uncertainty are not readily accessible.
Hence, we define ${\bf R}_{t}$ as
\begin{align}
  {\bf R}_{t} = \left[ \begin{matrix}
    \sigma_{p}^{2} \tilde{{\bf H}}_{t}^{-1} & {\bf 0} & {\bf 0} & {\bf 0} \\
    {\bf 0} & \sigma_{v}^{2} {\bf I} & {\bf 0} & {\bf 0} \\
    {\bf 0} & {\bf 0} & \sigma_{\omega}^{2} {\bf I} & {\bf 0} \\
    {\bf 0} & {\bf 0} & {\bf 0} & \sigma_{a}^{2} {\bf I} \\
  \end{matrix} \right],
\end{align}
where $\sigma^{2}$ represents positive arbitrary constants used to control the noise level. These values are determined through experiments.

\subsection{Keyframe selection and local map update}

After completing the scan matching process, keyframe selection is carried out.
The newly obtained scan matching result ${}^{\rm w}\tilde{{\bf T}}_{t}$ along with the corresponding deskewed scan points $\hat{ \mathcal{P} }_{t}$ are added to the keyframe set $\mathcal{K}$ if the correspondence rate $\gamma_{t}$ falls below a predefined threshold $\gamma_{\rm th}$.

Should a new keyframe be incorporated, the local map undergoes an update.
$N$ keyframes nearest to ${}^{\rm w}\tilde{{\bf p}}_{t}$ are identified within $\mathcal{K}$, and their associated scan points are utilized for the reconstruction of the local map.
It should be noted that this method of building the local map might lead to misestimations in odometry due to inconsistencies in the local map, which can arise from the accumulated error of odometry.
However, it plays a significant role in closing loops, which is crucial for accurate mapping and localization over longer periods.
Thus, the approach to building the local map should be adjusted according to the specifics of the target environments to balance between immediate odometry accuracy and long-term loop closure benefits.

\section{Experiments}
\label{sec:experiments}

\subsection{Setup}

To evaluate the performance of KLIO, we utilized the Stereo-Cam dataset from the Newer College dataset~\cite{ramezani2020newer}.
This dataset includes measurements from an Ouster OS-1 (Gen 1) 64 beam LiDAR equipped with an embedded IMU.
We employed its LiDAR and IMU data for our analysis.
Additionally, the dataset provides ground truth poses representing the trajectory of the left camera.
Therefore, we estimated its pose and assessed the Absolute Pose Error (APE) between our estimates and the ground truth using evo~\cite{grupp2017evo}.
For the evaluation, the LIO estimated the trajectory from the origin, and its trajectory was aligned with the ground truth trajectory using the $SE(3)$ Umeyama alignment method implemented in evo.
The parameters utilized in this study for KLIO are shown in Table~\ref{tab:parameters}.

\begin{table}[!t]
\begin{center}
\caption{The parameters used in KLIO.}
\begin{tabular}{|c|c|c|c|c|} \hline
$q = 1$                & $r = 0.1~[{\rm m}]$      & $\epsilon = 0.5~[{\rm m}]$    & $\gamma_{\rm th} = 0.8$ & $N = 20$                           \\ \hline
$\sigma_{p}^{2} = 100$ & $\sigma_{v}^{2} = 0.1$ & $\sigma_{\omega}^{2} = 0.1$ & $\sigma_{a}^{2} = 0.1$  & $\boldsymbol \Sigma_{0} = {\bf I}$ \\ \hline
\end{tabular}
\label{tab:parameters}
\end{center}
\end{table}

For comparative analysis, we utilized three methods: FAST-LIO2~\cite{FAST-LIO2}, Direct LIO (DLIO)\cite{Direct-LIO}, and our self-implemented hierarchical-geometric-observer-based LIO (GEO)\cite{Lopez2023arXiv}.
The study in~\cite{Direct-LIO} presents APEs on the Stereo-Cam dataset, highlighting FAST-LIO2 as the superior method among tightly-coupled methods.
Furthermore, DLIO was demonstrated to outperform all other methods, both tightly- and loosely-coupled, according to~\cite{Direct-LIO}.
Therefore, FAST-LIO2 and DLIO were chosen for comparison.
Our method bears similarity to DLIO; however, it differs in the update process.
Additionally, the scan matching processes between our method and DLIO also exhibit slight differences.
To ensure a fair comparison of the update performances between EKF and GEO, we implemented the update process using GEO.
The parameters used for the GEO-based update were set to match those shown in the GitHub repository~\footnote{ \url{https://github.com/vectr-ucla/direct_lidar_inertial_odometry} }.
It is important to note that we re-conducted experiments with FAST-LIO2 and DLIO on the Stereo-Cam dataset ourselves, as we were unable to replicate the exact results reported in~\cite{Direct-LIO}.

\subsection{Results}

Table~\ref{tab:results} provides a summary of APE for each method across five sequences.
It is noteworthy that the estimates by FAST-LIO2 experienced slippage at the starting points of sequences 01 and 05.
The values presented with and without parentheses correspond to APEs excluding and including the slippage, respectively.

\begin{table*}[!t]
\begin{center}
\caption{Root-mean-square errors of absolute pose error in meter.}
\begin{tabular}{|c|c|c|c|c|c|}
\hline
                                                                       & 01\_short\_experiment                                   & 02\_long\_experiment & 05\_quad\_with\_dynamics                                & 06\_dynamic\_spinning & 07\_parkland\_mound \\ \hline
KLIO (ours)                                                            & 0.390                                                   & 0.368                & 0.101                                                   & 0.108                 & 0.157               \\ \hline
GEO                                                                    & 0.451                                                   & 0.395                & {\bf 0.100}                                                   & 0.116                 & 0.157               \\ \hline
DLIO \cite{Direct-LIO}                                                 & {\bf 0.358}                                                   & 0.403                & 0.101                                                   & 0.109                 & 0.143               \\ \hline
\begin{tabular}[c]{@{}c@{}}FAST LIO2 \cite{FAST-LIO2} \\ (exclude slippage)\end{tabular} & \begin{tabular}[c]{@{}c@{}}0.393\\ (0.355)\end{tabular} & {\bf 0.365}                & \begin{tabular}[c]{@{}c@{}}0.316\\ (0.098)\end{tabular} & {\bf 0.084}                 & {\bf 0.136}               \\ \hline
\end{tabular}
\label{tab:results}
\end{center}
\end{table*}

All methods achieved accurate estimations.
The accuracy of FAST-LIO2 surpassed that of the other methods, especially when excluding the instances of slippage, which were notably precise.
These findings indicate that FAST-LIO2 was the most accurate method among those tested.
However, it's important to note that FAST-LIO2 experienced slippage, and its APE in sequence 05 was less accurate.
This issue is attributed to the method's feature extraction process, as the field of view of the LiDAR at the starting points of sequences 01 and 05 was limited.

\begin{figure}[!t]
  \begin{center}
    \includegraphics[width = 85 mm]{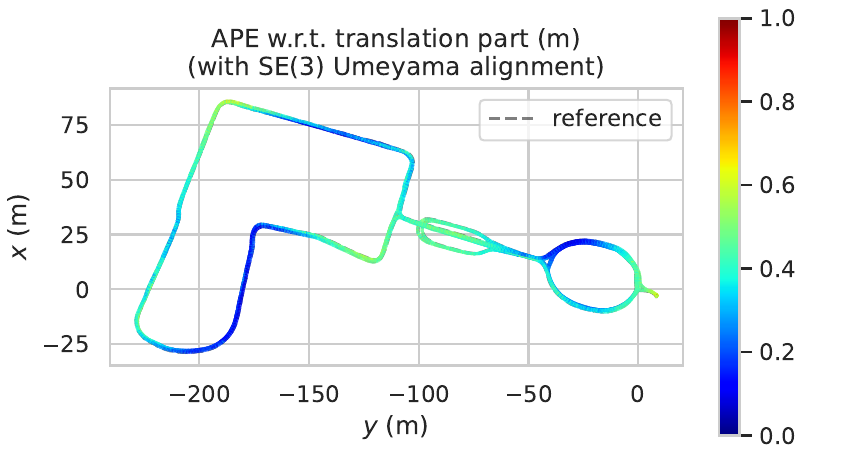}
    \caption{The evaluation of absolute pose error (APE) using evo~\cite{grupp2017evo} against the estimates made by KLIO on the 02\_long\_experiment sequence.}
    \label{fig:evo_long}
  \end{center}
\end{figure}

KLIO, GEO, and DLIO all performed accurate tracking without experiencing slippage or any other significant issues, making it challenging to definitively determine the best method among them.
However, KLIO generally achieved higher accuracy compared to GEO.
Therefore, we conclude that our proposed update process, which is based on EKF, offers better performance than the update process based on the geometric observer.
Figure~\ref{fig:evo_long} shows the evaluation of APE using evo against the estimates made by KLIO on the 02\_long\_experiment sequence.
Note that KLIO did not achieve the best performance; however, its mapping accuracy was quite high, as shown in Fig.~\ref{fig:result_ncd_long}.

\subsection{Qualitative results}

\begin{figure}[!t]
  \begin{center}
    \includegraphics[width = 85 mm]{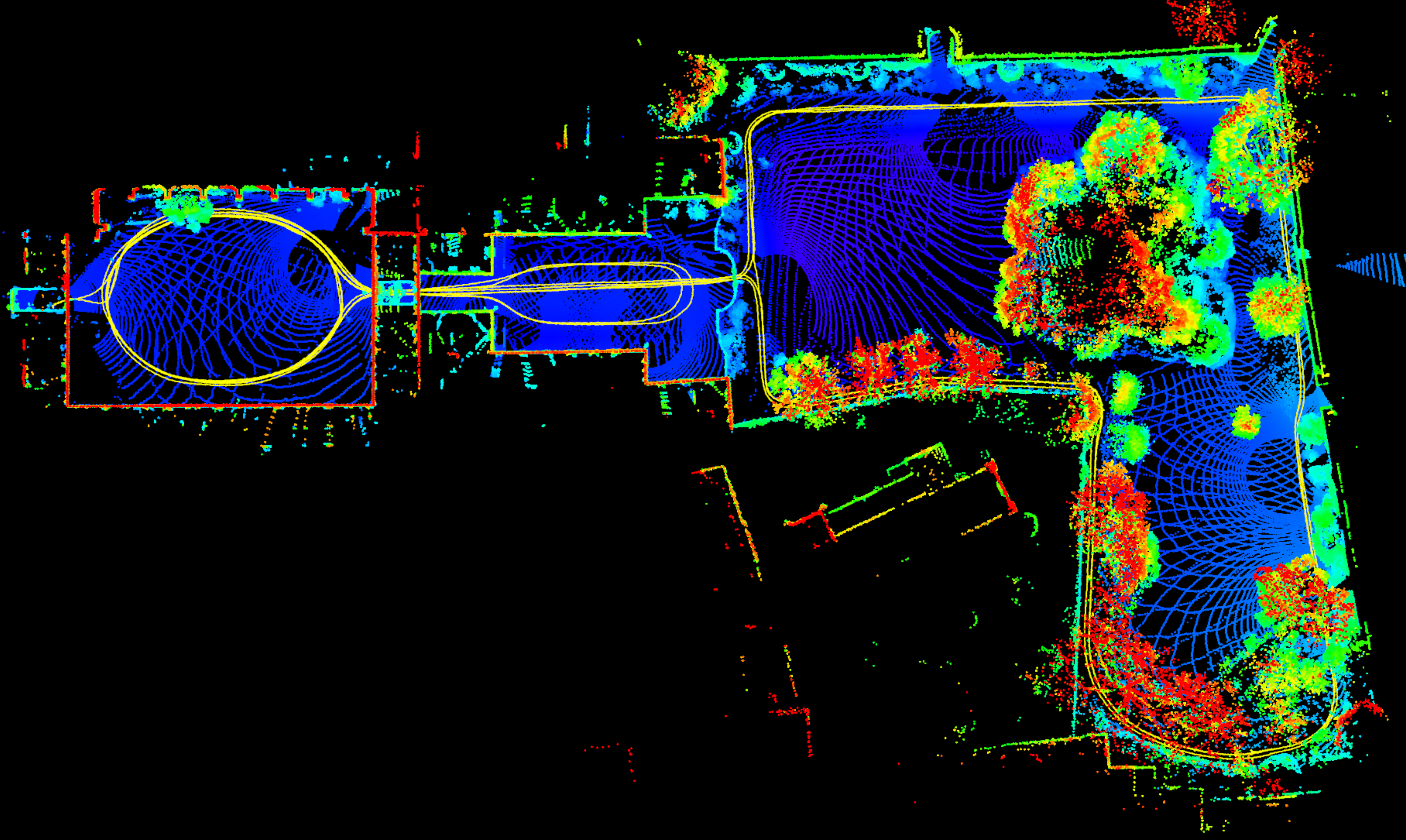}
    \caption{The local map construction on the 01\_short\_experiment sequence with $\gamma_{\rm th}$ set to 0.5, resulting in the registration of only 15 keyframes. The points color indicates height from the origin.}
    \label{fig:long_local_map}
  \end{center}
\end{figure}

We further conducted some experiments to qualitatively evaluate KLIO.
Figure~\ref{fig:long_local_map} illustrates the construction of the local map for the 01\_short\_experiment sequence with $\gamma_{\rm th}$ set to 0.5.
Despite the registration of only 15 keyframes, KLIO was able to achieve accurate mapping.
This outcome indicates that KLIO can operate efficiently by adjusting the threshold $\gamma_{\rm th}$ for point correspondences, demonstrating its capability to maintain mapping accuracy with a selective approach to keyframe registration.

\begin{figure}[!t]
  \begin{center}
    \includegraphics[width = 80 mm]{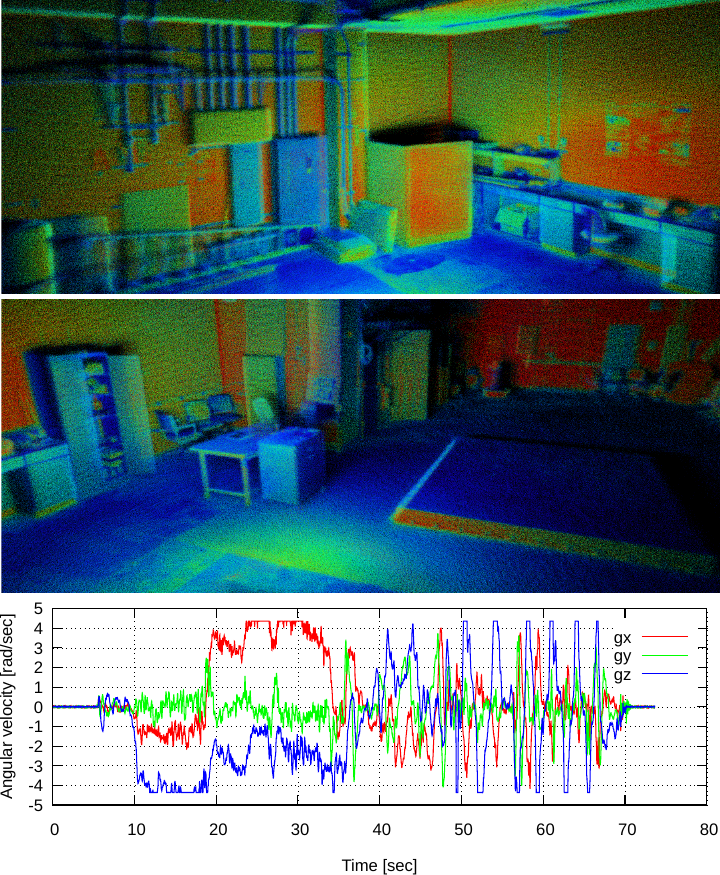}
    \caption{Mapping results using KLIO on data that involved aggressive spinning, where the maximum angular velocities measured by the IMU exceeded $4~{\rm rad/sec}$. The colors represent intensity levels, with blue and red indicating low and high values.}
    \label{fig:aggressive_spinning}
  \end{center}
\end{figure}

Additionally, we conducted an experiment using data that involved aggressive spinning, where the maximum angular velocities measured by the IMU exceeded $4~{\rm rad/sec}$\footnote{ \url{https://drive.google.com/file/d/1Sp_Mph4rekXKY2euxYxv6SD6WIzB-wVU/view} }.
Despite the challenging conditions, KLIO was able to accurately map the environment as shown in Fig.~\ref{fig:aggressive_spinning}.
This outcome underscores the superior performance and robustness of KLIO, highlighting its capability to maintain accurate mapping and localization under highly aggressive movement.

\section{Conclusion}
\label{sec:conclusion}

This paper has presented a LiDAR-inertial odometry (LIO) method based on the Extended Kalman Filter (EKF), which we refer it to as KLIO.
Contrary to the majority of recent LIO methods that are grounded in optimization-based approaches, our approach demonstrates that the LIO framework can be formulated through recursive Bayesian filtering.
The architecture of KLIO is streamlined, incorporating scan matching processes between LiDAR and local point clouds, complemented by an update mechanism that is informed by the results of scan matching.
In the experiments, we used the Stereo-Cam dataset from the Newer College dataset and compared KLIO with three methods, including the state-of-the-art in both tightly- and loosely-coupled approaches.
The evaluation was performed in terms of the absolute pose error and the results indicated that the performance of KLIO is comparable to that of state-of-the-art methods.

In \cite{AkaiJFR2023}, we have introduced a reliability estimation method for LiDAR-based localization based on recursive Bayesian filtering for realizing reliable autonomous systems.
We also would like to achieve reliability estimation of LIO in the future work.





%

\section*{ACKNOWLEDGMENT}

This work was supported by KAKENHI under Grant 23K03773.

\balance
\bibliographystyle{unsrt}
\bibliography{reference.bib}


\end{document}